\documentclass[letterpaper, 10 pt, conference]{ieeeconf}  

\IEEEoverridecommandlockouts                              %
\usepackage{graphicx}
\usepackage{amsmath}
\usepackage{amssymb}
\usepackage{booktabs}
\usepackage{flushend}
\usepackage{svg}
\usepackage{makecell}
\usepackage{multirow}
\usepackage{subcaption}

\usepackage{hyperref}
\hypersetup{colorlinks,allcolors=black}

\usepackage[capitalize]{cleveref}
\crefname{section}{Sec.}{Secs.}
\Crefname{section}{Section}{Sections}
\Crefname{table}{Table}{Tables}
\crefname{table}{Tab.}{Tabs.}

\begin{document}

\title{Robust Scene Change Detection Using Visual Foundation Models and Cross-Attention Mechanisms}

\author{Chun-Jung Lin\qquad
Sourav Garg\qquad
Tat-Jun Chin\qquad
Feras Dayoub\\\\
Australian Institute for Machine Learning (AIML), University of Adelaide, Australia\\
{\tt\small\{chun-jung.lin, sourav.garg, tat-jun.chin, feras.dayoub\}@adelaide.edu.au}
}

\maketitle

\begin{abstract}
We present a novel method for scene change detection that leverages the robust feature extraction capabilities of a visual foundational model, DINOv2, and integrates full-image cross-attention to address key challenges such as varying lighting, seasonal variations, and viewpoint differences. In order to effectively learn correspondences and mis-correspondences between an image pair for the change detection task, we propose to a) ``freeze'' the backbone in order to retain the generality of dense foundation features, and b) employ ``full-image'' cross-attention to better tackle the viewpoint variations between the image pair. We evaluate our approach on two benchmark datasets, VL-CMU-CD and PSCD, along with their viewpoint-varied versions. Our experiments demonstrate significant improvements in F1-score, particularly in scenarios involving geometric changes between image pairs. The results indicate our method's superior generalization capabilities over existing state-of-the-art approaches, showing robustness against photometric and geometric variations as well as better overall generalization when fine-tuned to adapt to new environments. Detailed ablation studies further validate the contributions of each component in our architecture. Our source code is available at: https://github.com/ChadLin9596/Robust-Scene-Change-Detection.

\end{abstract}

\section{Introduction}
\label{sec:intro}

Scene change detection (SCD) is a crucial capability for autonomous robotic systems, enabling applications such as autonomous navigation, real-time map update, environmental monitoring, and infrastructure inspection. By identifying differences between images captured at different times, SCD can provide essential insights for maintaining up-to-date maps~\cite{vl-cmu-cd, CDreview}, monitoring environmental changes~\cite{taneja2011image}, and ensuring security~\cite{underwood2013explicit}. 

Despite its importance, scene change detection poses significant challenges due to various factors such as lighting variations, seasonal variations, and viewpoint differences, which can lead to false positives and negatives, thus compromising detection reliability.

Over the years, various approaches have been developed to tackle SCD, from traditional image processing techniques to sophisticated deep learning models. Traditional methods, such as image differencing and optical flow techniques, often struggle with complex scenarios involving photometric and geometric changes. Deep learning has significantly advanced the field, enabling the extraction and integration of powerful features from images. Notable approaches include Fully Convolutional Networks (FCNs)~\cite{long2015fully} and Siamese Networks~\cite{daudt2018fully}, which have demonstrated improved performance in change detection tasks by leveraging hierarchical feature representations and the ability to compare image pairs effectively~\cite{drtanet, sscdnet, ChangeNet}.

\begin{figure}[t]
    \centering
    \includegraphics[width=0.47\textwidth]{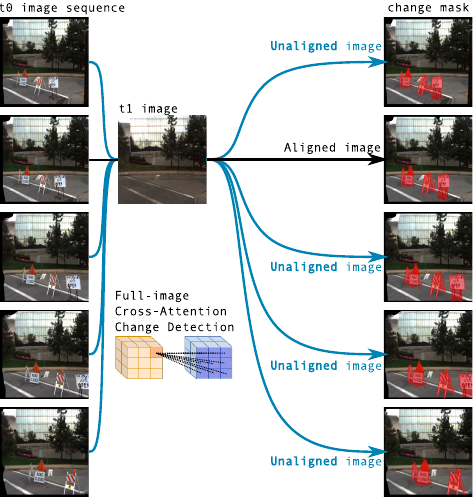}
    \caption{\textbf{\textit{Unaligned images change detection:}} we approach the change detection problem with \textit{cross attention module}, making robust detection on unaligned scenes.}
    \label{fig:hook-figure}
\end{figure}

Given the limitations of current methods, we propose a more robust approach to change detection. Our method addresses these gaps by leveraging a visual foundational model as the backbone network for its robust feature extraction capabilities and integrating cross-attention to register the features. This combination allows for accurate handling of correspondences and mis-correspondences between image pairs, effectively mitigating the impact of photometric and viewpoint changes, and leading to better generalization, as shown in \cref{fig:hook-figure}.

Our key contributions are as follows:

\begin{itemize}
    \item We propose a novel approach to scene change detection (SCD) that leverages the robust feature extraction capabilities of a visual foundational model.
    \item We demonstrate the use of a full-image cross-attention mechanism to effectively address viewpoint variations between image pairs.
    \item We perform extensive evaluations on the VL-CMU-CD~\cite{vl-cmu-cd} and PSCD~\cite{sscdnet} datasets, including newly created viewpoint-varied versions.
    \item We conduct detailed ablation studies to validate the effectiveness of each architectural component and provide insights into the contributions of our design choices.
\end{itemize}

\begin{figure*}[t]
    \centering
    \includegraphics[width=0.95\textwidth]{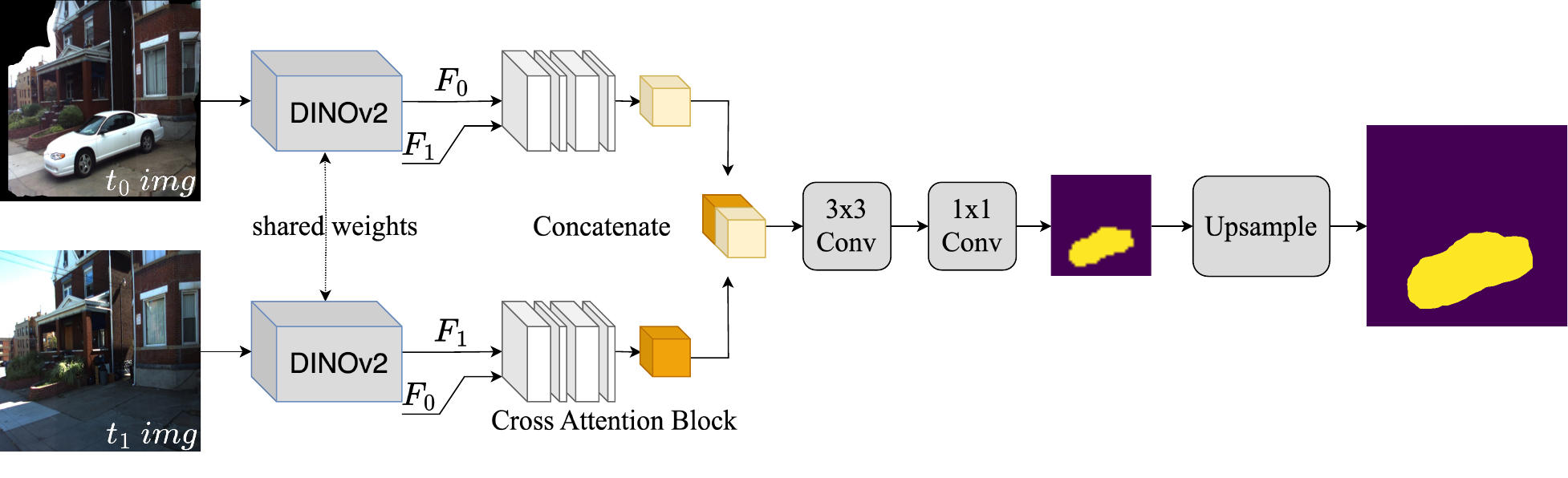}
    \caption{\textbf{\textit{Architecture:}} An overview of the proposed change detection architecture, where the backbone is kept frozen to achieve better overall generalization. $F_0$ and $F_1$ are the dense feature from $t_0$ and $t_1$ images, respectively.}
    \label{fig:architecture}
\end{figure*}

\section{Related Works}
\label{sec: relwork}

Various scene change detection (SCD) approaches have been developed, ranging from traditional image differencing techniques to more advanced deep learning-based methods, each addressing different aspects of the problem. Traditional approaches, such as image differencing and optical flow~\cite{CDreview}, often struggle to handle complex scene variations, particularly under changing lighting conditions and geometric transformations. 

Deep learning has significantly advanced SCD by leveraging powerful feature representations. Some methods focus on detecting changes in 2D images~\cite{drtanet, sscdnet, c3po, wang2021transcd}, while others target 3D data~\cite{krawciw2023change, plachetka2023dnn, yew2021city} or combine both 2D and 3D information~\cite{he2022diff, nagy2021changegan}. Given the time-consuming nature of collecting real-world datasets, synthetic datasets like ChangeSim~\cite{park2021changesim}, COCO-Inpainted~\cite{cyws}, Kubric-Change~\cite{cyws}, and KC-3D~\cite{cyws-3d} are often used to supplement training data in SCD research.

The application context of SCD methods varies significantly. Ground, satellite, and aerial imagery are widely used for change detection in remote sensing~\cite{rs16132355}, focusing on large-scale environmental monitoring. In contrast, street-view images are commonly employed in autonomous vehicle applications~\cite{vl-cmu-cd}, where accurate and timely detection of scene changes is crucial for navigation and safety.

Some methods not only detect changes but also classify and recognize specific types of changes. For example, C3PO~\cite{c3po} and ChangeSim~\cite{park2021changesim} categorize changes into appearance, disappearance, or object exchange, providing detailed information about the nature of the changes. Similarly, SSCDNet~\cite{sscdnet} incorporates semantic segmentation to recognize different types of changes, integrating object-level understanding into the change detection process.

A major challenge in SCD, especially for robotics, is handling viewpoint variations. Researchers often insert feature comparators between encoders and decoders to register features across different aligned images. For instance, SSCDNet~\cite{sscdnet} employs correlation layers to address viewpoint differences by establishing feature registration, while C3PO~\cite{c3po} proposes multiple subtraction branches to classify changes by learning each type of change separately. Alternatively, some methods treat change detection and feature registration as independent tasks, using optical flow labels to assist in change detection~\cite{lee2024semi, park2022dual}.

Inspired by advancements in natural language processing, attention mechanisms have been incorporated into SCD to improve feature alignment. Self-attention is utilized in TransCD~\cite{wang2021transcd} for scene change detection, and DR-TANet~\cite{drtanet} leverages attention layers to address correlation challenges in change tracking. Co-attention mechanisms are employed in CYWS~\cite{cyws} to register features while predicting bounding boxes for changed objects. Beyond change detection, attention mechanisms have been used to register features across different domains; for example, attention layers help align street-view images with satellite imagery for localization tasks~\cite{yuan2024cross}, and register images with varying styles, locations, and orientations~\cite{wiles2021co}. 

Compared to methods that classify or recognize the semantic meaning of changes, our approach is orthogonal in its focus on robustly detecting changes under significant viewpoint variations. We leverage a visual foundational model for robust feature extraction and introduce a full-image cross-attention mechanism to effectively handle viewpoint differences between image pairs. By freezing the backbone network during training, we retain the generality of dense foundational features, enhancing the reliability of change detection. Our method is complementary to classification-based approaches and can be integrated with them to address both geometric and semantic challenges in scene change detection. 

\section{Problem Statement}
\label{sec: define}

\textit{Our objective is to segment an outdated image into changed and unchanged regions by comparing it with a new image regardless of whether images are pixel-aligned or not.} The primary challenges include differences in camera angles and positions, resulting in geometric transformations that render direct pixel-wise comparison ineffective. Perfectly aligned image pairs are rare in real-world applications, making pixel-wise alignment difficult.

\section{Methodology}
\label{sec: method}

We follow the conventional strategies~\cite{drtanet, sscdnet, c3po} that obtain dense feature $F_0\ \&\ F_1$ from a CNN-based encoder for each image in an image pair. Different from the backbone ResNet-18~\cite{resnet} and VGG-16~\cite{vgg} used in~\cite{drtanet, sscdnet, c3po}, we select and freeze the smallest DINOv2~\cite{dinov2} as our backbone for its visual foundational ability. Next, the learnable cross-attention modules are employed to find correspondence and mis-correspondences between the dense feature $F_0\ \&\ F_1$. With the correspondence signals from each image in a pair extracted, we concatenate them and perform a series of 2D convolution layers as the decoder to predict a change mask, as shown in~\cref{fig:architecture}.

\subsection{Image Encoding:}

The DINOv2~\cite{dinov2} backbone is designed to produce all-purpose visual features and is constructed based on the Vision Transformer (ViT) model~\cite{vit}. Because of its strong visual representation capabilities, we do not train or fine-tune it to our datasets. Instead, we freeze the smallest one (21M parameters) and use it to develop this model.

The frozen backbone will generate rich features in every image patch (14$\times$14 pixels). Thus, an image $\in\mathbb{R}^{H\times W \times 3}$ will be transformed into a dense feature $F\in\mathbb{R}^{h\times w\times f}$, where $H$ equals $14 \times h$ and $W$ equals $14 \times w$.

\subsection{Image Comparator:}

We use cross-attention modules for the change detection task, as they can register features between pixels from different images, even when not pixel-wise aligned~\cite{yuan2024cross}. The cross-attention module acts as our image comparator, registering correspondences and mis-correspondences between two images. Specifically, two cross-attention blocks are formed to learn and extract signals from $F_0$ and $F_1$ by given $F_1$ and $F_0$, respectively. Later, these signals are concatenated to form an advanced feature matrix $\in\mathbb{R}^{h\times w\times 2f}$ and fed into the segmentation head to decode and generate a prediction mask. To further identify the ability of the cross-attention module, we compare different image comparators at \cref{subsec:ablation}.

\subsection{Change Mask Prediction:}

First, a 3$\times$3 convolution layer is formed to halve the number of features from the advanced feature matrix extracted by the image comparator. Second, for segmentation prediction, a 1x1 convolution layer is formed to decode the dense signals into two channels: change and unchanged. Lastly, an upsampling layer is applied to upsample the predicted mask $\in\mathbb{R}^{h\times w\times 2}$ to the target prediction $\in\mathbb{R}^{H \times W \times 2}$.

Following the setting of C3PO~\cite{c3po}, we use the weighted softmax cross-entropy loss function for the segmentation prediction. The prediction will be two classes: change and unchanged.

\section{Experiments}
\label{sec: exp}

\setlength{\tabcolsep}{2pt}
\begin{table}[htbp]
    \raggedleft
    \caption{\textbf{\textit{Change detection datasets:}} we list the number of image pairs, the number of scenes/sources, and environments for data choices. The ``imgs" and ``env." represent ``images" and ``environment", respectively.}
    \resizebox{0.5\textwidth}{!}{
        \begin{tabular}{l | c | c | c }
            \toprule
            Dataset & \# of pairs & sources & real env.? \\
            \midrule
            CDnet2012~\cite{CDnet} & 90,000 & 31 videos & Yes, outdoor\\
            CDnet2014~\cite{cdnet2014} & 70,000 & 22 videos & Yes,  outdoor \\
            ChangeSim~\cite{park2021changesim} & 130,000 & 80 videos (10 scenes) & No, indoor \\
            \textbf{VL-CMU-CD}~\cite{vl-cmu-cd} & 1,362 & \textbf{152 sequences} & Yes, outdoor \\ 
            \textbf{PSCD}~\cite{sscdnet} & 11,550 & \textbf{770 panoramic imgs.} & Yes, outdoor \\
            \bottomrule
        \end{tabular}
    }
    \label{tab:table-data-1}
\end{table}

\begin{table}[htbp]
    \raggedleft
    \caption{\textbf{\textit{Aligned and Unaligned Test sets:}} the definition and number of image pairs of each test set.}
    \resizebox{0.49\textwidth}{!}{
        \begin{tabular}{l | c | c | c }
            \toprule
            Test Set & augmentation & \# of pairs & comments\\
            \midrule
            \multirow{3}{*}{\makecell[l]{VL-CMU-CD~\cite{vl-cmu-cd} \\ ($504\times504$)}}
                       & original & 429 & Coarsely aligned \\ 
                       & Diff-1 & 375*2 & adjacent pairs (distance 1) \\
                       & Diff-2 & 323*2 & adjacent pairs (distance 2) \\
            \midrule
            \multirow{3}{*}{\makecell[l]{PSCD~\cite{sscdnet} \\ ($224\times224$)}}
                        & original & 1,155 & Coarsely aligned \\ 
                        & Diff-1 & 1,078*2 & adjacent pairs (distance 1) \\
                        & Diff-2 & 1,001*2 & adjacent pairs (distance 2) \\
            \bottomrule
        \end{tabular}
    }
    \label{tab:table-data-2}
\end{table}
\setlength{\tabcolsep}{6pt}

\begin{table*}[t]
    \centering
    \caption{\textbf{\textit{F1-score after training on VL-CMU-CD:}} we compare different backbones, aligned/unaligned datasets, and inference time. Among all baselines, our method with the DinoV2 backbone achieves the best results on aligned/unaligned datasets. The results in ``Inference" column are average of inferring 10,000 images. The ``Avg." represents the average metric.}
    \resizebox{0.85\textwidth}{!}{
        \begin{tabular}{l | c | c c c c | c | c | c}
            
            \toprule
            
             & & 
            \multicolumn{4}{c|}{Vl-CMU-CD~\cite{vl-cmu-cd}} &
            \multicolumn{1}{c|}{PSCD~\cite{sscdnet}} &
            \multicolumn{1}{c}{both} & 
            \multicolumn{1}{c}{Inference} \\
            Method & Backbone & Aligned & Diff-1 & Diff-2 & Avg. & Aligned & Avg. & Time (ms) \\
            
            \midrule
            
            TransCD~\cite{wang2021transcd} & Resnet-18 & 0.558 & 0.487 & 0.454 & 0.492 & - & - & 4.48\\
            DR-TANet~\cite{drtanet} & Resnet-18 & 0.607 & 0.577 & 0.569 & 0.581 & 0.023 & 0.365 & 6.79\\
            CDNet~\cite{sakurada2017dense} & U-net & 0.675 & 0.613 & 0.601 & 0.623 & - & - & 5.53\\
            CSCDNet~\cite{sscdnet} & Resnet-18 & 0.766 & - & - & - & - & - & - \\
            C-3PO~\cite{c3po} & Resnet-18 & \textbf{0.795} & 0.721 & 0.693 & 0.728 & 0.048 & 0.465 & 5.02\\
            
            \midrule
            
            \textbf{ours} & Resnet-18 & 0.687 & 0.679 & 0.672 & 0.679 & 0.097 & 0.453 & 3.82 \\
            \textbf{ours} & DinoV2 & \textbf{0.795} & \textbf{0.760} & \textbf{0.739} & \textbf{0.761} & \textbf{0.337} & \textbf{0.597} & 6.64 \\
            
            \bottomrule
            
        \end{tabular}
    }
    \label{tab:table-backbone}
\end{table*}

\subsection{Datasets}

Many change detection datasets are publicly available for benchmarking. CDnet2012~\cite{CDnet} and CDnet2014~\cite{cdnet2014} released a series of videos to detect changes in outdoor CCTV cameras. ChangeSim~\cite{park2021changesim} recorded drone videos in simulated warehouses to identify artificial changes. These datasets provide enormous images but are limited to a few scenes. VL-CMU-CD~\cite{vl-cmu-cd} is a dataset aiming to update large-scale autonomous vehicle navigation maps and provide many more city scenes, making it a challenging change detection dataset. PSCD~\cite{sscdnet} provides hundreds of panoramic image pairs for semantic change detection tasks in different locations. Both VL-CMU-CD and PSCD have fewer image pairs compared to CDnet2012, CDnet2014 and ChangeSim, but they contain more diverse scenes for evaluating change detection methods. Thus, we choose VL-CMU-CD and PSCD to evaluate our methods. The numbers of image pairs and scenes from all datasets are listed in \cref{tab:table-data-1}.

\paragraph{VL-CMU-CD:}

The VL-CMU-CD dataset consists of 933 coarsely aligned image pairs in the training set and 429 in the test set. Following C3PO's work~\cite{c3po}, the training set is augmented to 3,732 pairs by rotation. Additionally, we split 408 pairs from the training set as the validation set, making 3,324 pairs for training, 408 pairs for validation, and 429 pairs for testing.

\paragraph{PSCD:}

Following the work~\cite{sscdnet}, we crop each panoramic image to 15 images, making 11,550 aligned image pairs from 770 panoramic image pairs. We further divide pairs into 9,240 for training, 1,155 for validation, and 1,155 for testing.

\paragraph{Unaligned scenes from aligned scenes:}

Street-view images captured at different timestamps often exhibit geometric transformations. To make the datasets more challenging and close to real utilization, we create unaligned datasets from VL-CMU-CD and PSCD. Specifically, we make new image pairs by adjacent neighbors from the same sequence of VL-CMU-CD and the same panoramic image of PSCD. \cref{tab:table-data-2} shows the number of image pairs of these unaligned datasets, which will be used to evaluate the performance of each approach. 

\subsection{Evaluation Metric:}

Following previous methods~\cite{vl-cmu-cd, drtanet, c3po}, we use the F1-score, the harmonic mean of precision and recall, as the evaluation metric. For each image pair in the VL-CMU-CD and PSCD, we compute the F1-score for a predicted change mask. Then, we average the scores from the test sets. 

\begin{table}[!t]
    \centering
    \caption{\textbf{\textit{Different Viewpoint Augmentation:}} we report F1-score on VL-CMU-CD dataset after training with the unaligned dataset.}
    \resizebox{0.5\textwidth}{!}{
        \begin{tabular}{l | c | c c c c}
            \toprule
            \multicolumn{1}{l|}{} &
            \multicolumn{1}{c|}{} & 
            \multicolumn{4}{c}{Vl-CMU-CD~\cite{vl-cmu-cd}} \\
            Method & \makecell[c]{Diff-View \\ Augment} & Aligned & Diff-1 & Diff-2 & Avg. \\
            \midrule
            DR-TANet~\cite{drtanet} & No & 0.607 & 0.577 & 0.569 & 0.581\\
            DR-TANet~\cite{drtanet} & Yes & 0.536 & 0.535 & 0.536 & 0.535\\
            \midrule
            CDNet~\cite{sakurada2017dense} & No & 0.675 & 0.613 & 0.601 & 0.623\\
            CDNet~\cite{sakurada2017dense} & Yes & 0.524 & 0.521 & 0.517 & 0.521\\
            \midrule
            C-3PO~\cite{c3po} & No & \textbf{0.795} & 0.721 & 0.693 & 0.728\\
            C-3PO~\cite{c3po} & Yes & 0.706 & 0.703 & 0.698 & 0.702\\
            \midrule
            \textbf{ours} & No & \textbf{0.795} & 0.760 & 0.739 & 0.761\\
            \textbf{ours} & Yes & 0.787 & \textbf{0.785} & \textbf{0.784} & \textbf{0.785}\\
            \bottomrule
        \end{tabular}
    }
    \label{tab:diff-view-augt}
\end{table}

\def\affinewidth{0.24\textwidth}
\def\affinewidthimg{\linewidth}

\begin{figure*}[t]

    \centering
    \begin{subfigure}[t]{\affinewidth}
        \centering
        \includegraphics[width=\affinewidthimg]{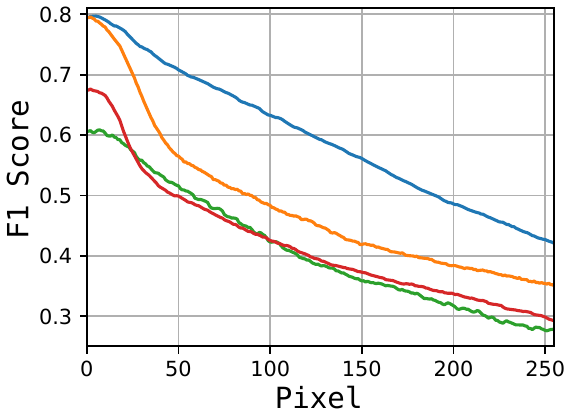}
        \caption{trans. wo. diff-view augment}
        \label{fig:affine-translate}
    \end{subfigure}%
    ~
    \begin{subfigure}[t]{\affinewidth}
        \centering
        \includegraphics[width=\affinewidthimg]{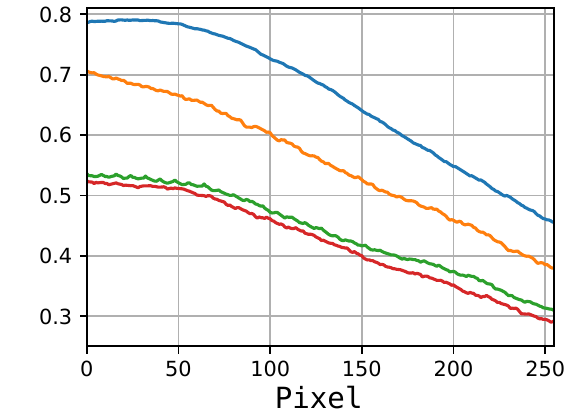}
        \caption{trans. w. diff-view augment}
        \label{fig:affine-rotate}
    \end{subfigure}%
    ~
    \begin{subfigure}[t]{\affinewidth}
        \centering
        \includegraphics[width=\affinewidthimg]{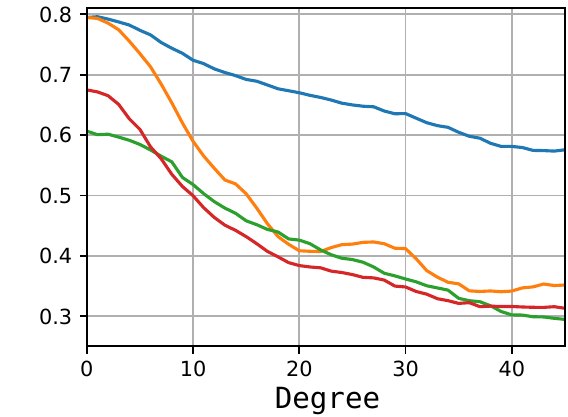}
        \caption{rot. wo. diff-view augment}
        \label{fig:subfig3}
    \end{subfigure}%
    ~
    \begin{subfigure}[t]{\affinewidth}
        \centering
        \includegraphics[width=\affinewidthimg]{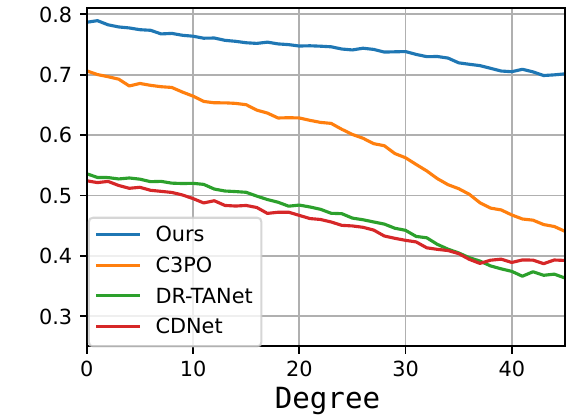}
        \caption{rot. w. diff-view augment}
        \label{fig:subfig4}
    \end{subfigure}
    \caption{\textbf{\textit{F1-score of Affine Transformation:}} we evalute F1-score after translate (trans.) or rotate (rot.) $t_0$ images from VL-CMU-CD test set. (a) and (b) are translation results without (wo.) and with (w.) different viewpoint augmentation (diff-view augment). (c) and (d) are rotation results before and after the augmentation. The blue line indicates ours results. C3PO, DR-TANet, and CDNet results are plotted as orange, green, and red lines, respectively.}
    \label{fig:fig-affine}
\end{figure*}

\def\imgwidth{0.09\textwidth}
\setlength{\tabcolsep}{2pt}

\begin{figure*}
    \centering
    \begin{tabular}{c c | c c | c c | c c | c c }
        $t_0$ img + gt. & $t_1$ img &
        \multicolumn{2}{c|}{ours} &
        \multicolumn{2}{c|}{C-3PO~\cite{c3po}} &
        \multicolumn{2}{c|}{CDNet~\cite{sakurada2017dense}} &
        \multicolumn{2}{c}{DR-TANet~\cite{drtanet}} \\ 

        & & wo. & w. & wo. & w. & wo. & w. & wo. & w.\\
        & & \multicolumn{2}{c|}{Diff-View-Augment} & \multicolumn{2}{c|}{Diff-View-Augment} & \multicolumn{2}{c|}{Diff-View-Augment} & \multicolumn{2}{c}{Diff-View-Augment} \\
        
        \toprule

         &
        \includegraphics[width=\imgwidth]{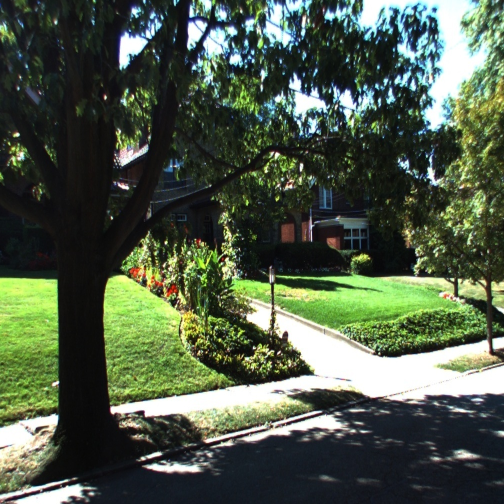} &
        \includegraphics[width=\imgwidth]{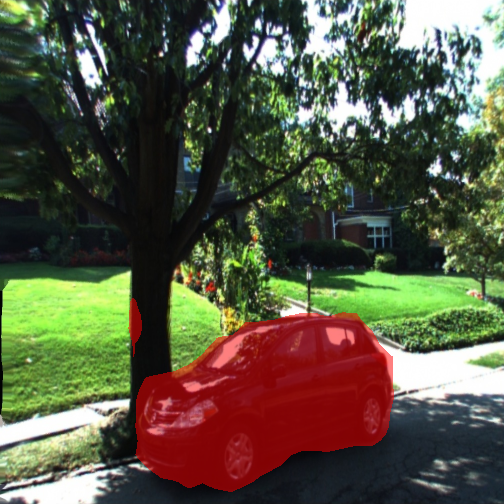} &
        \includegraphics[width=\imgwidth]{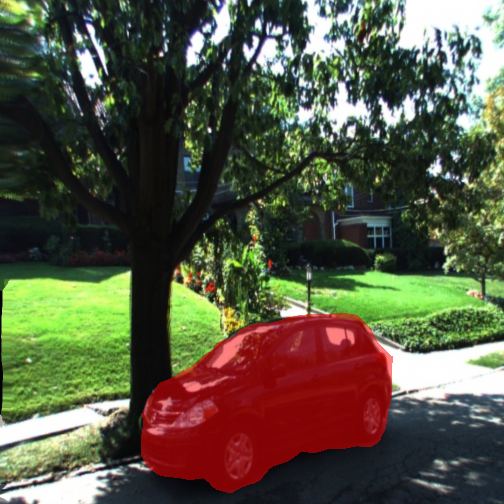} &
        \includegraphics[width=\imgwidth]{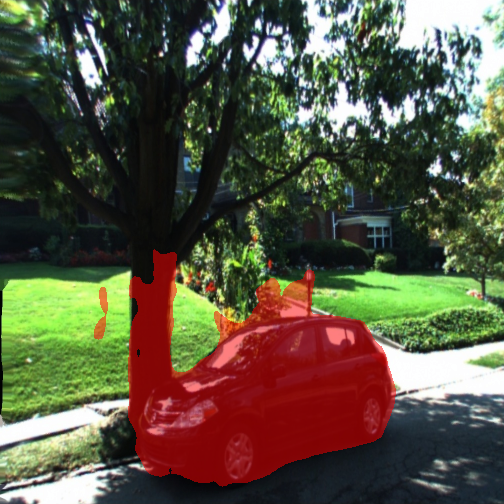} &
        \includegraphics[width=\imgwidth]{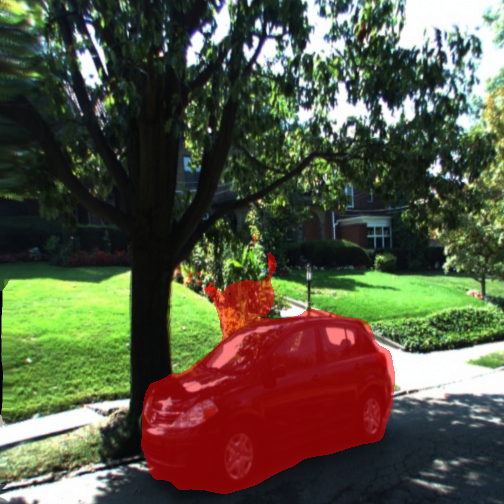} &
        \includegraphics[width=\imgwidth]{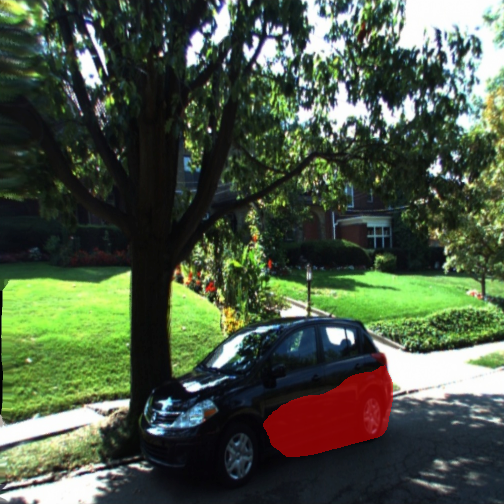} &
        \includegraphics[width=\imgwidth]{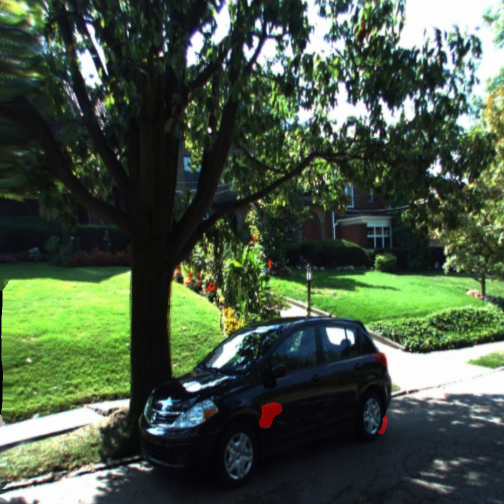} &
        \includegraphics[width=\imgwidth]{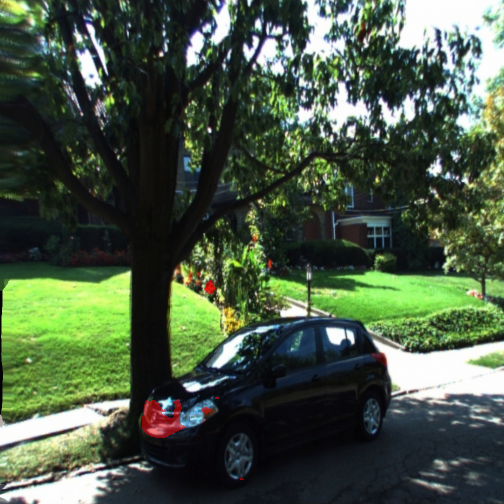} &
        \includegraphics[width=\imgwidth]{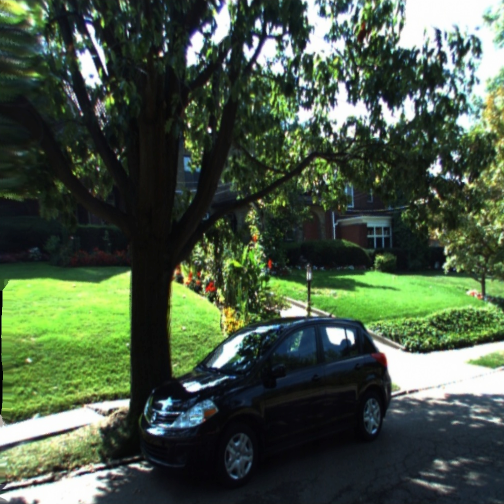} \\

        \includegraphics[width=\imgwidth]{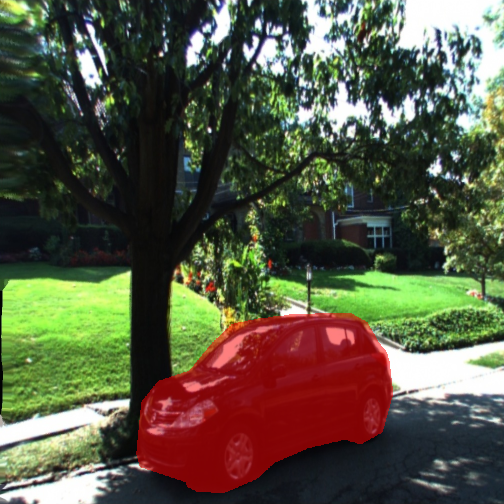} &
        \includegraphics[width=\imgwidth]{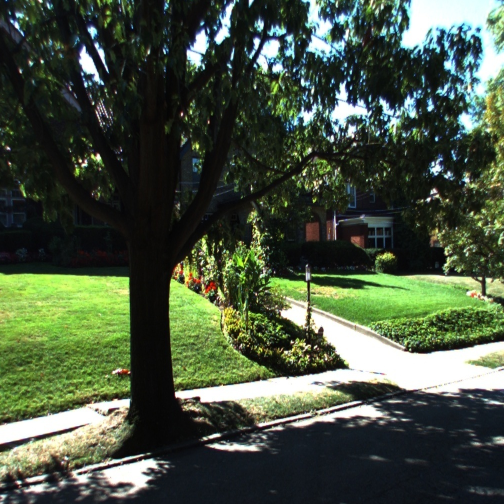} &
        \includegraphics[width=\imgwidth]{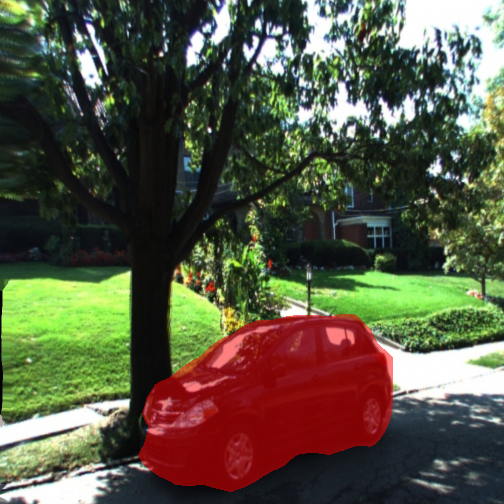} &
        \includegraphics[width=\imgwidth]{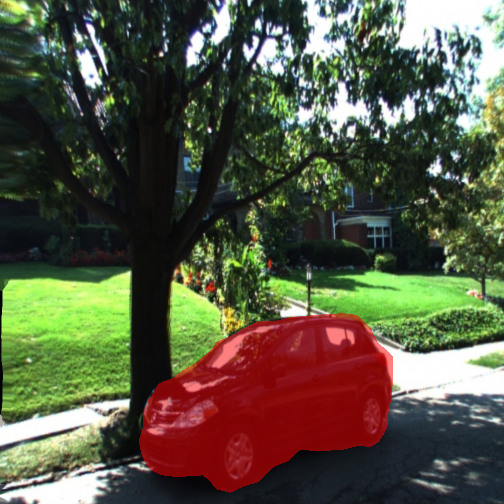} &
        \includegraphics[width=\imgwidth]{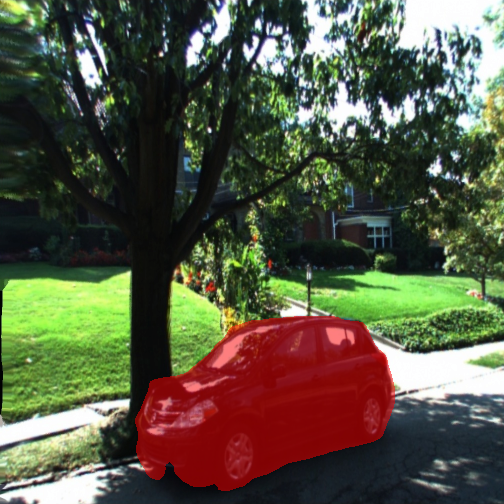} &
        \includegraphics[width=\imgwidth]{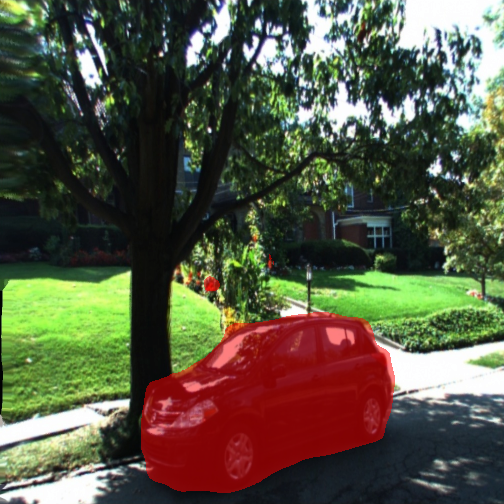} &
        \includegraphics[width=\imgwidth]{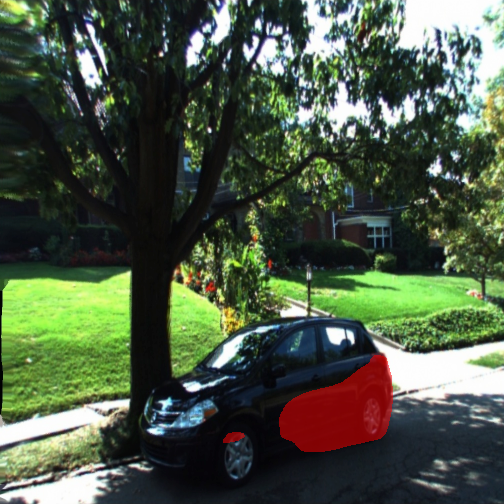} &
        \includegraphics[width=\imgwidth]{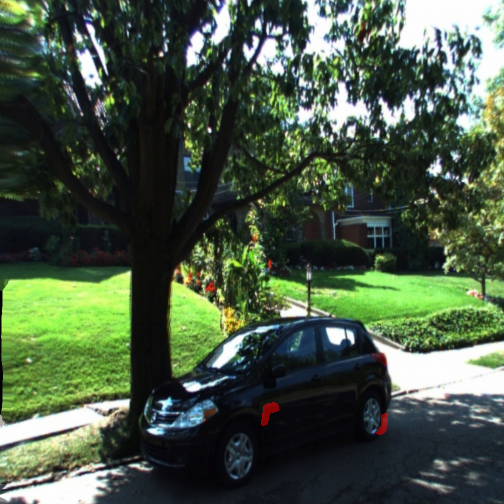} &
        \includegraphics[width=\imgwidth]{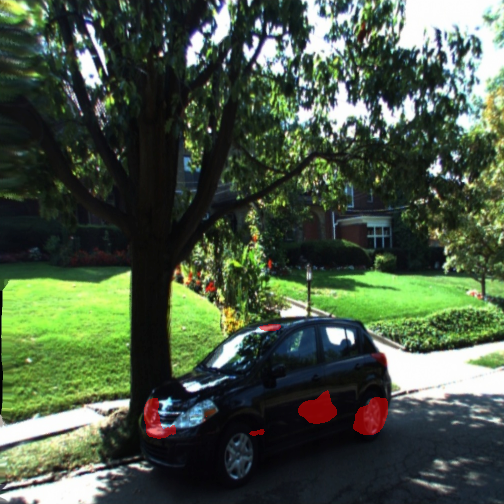} &
        \includegraphics[width=\imgwidth]{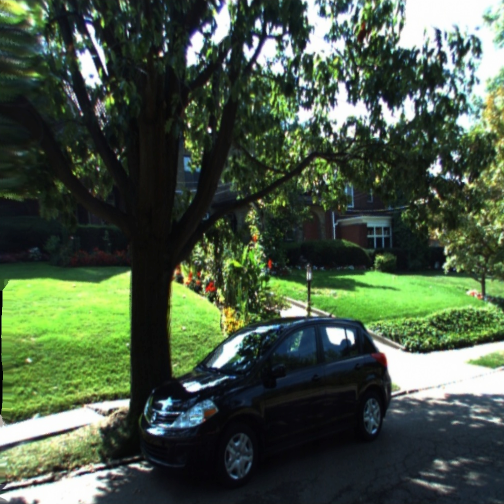} \\

         &
        \includegraphics[width=\imgwidth]{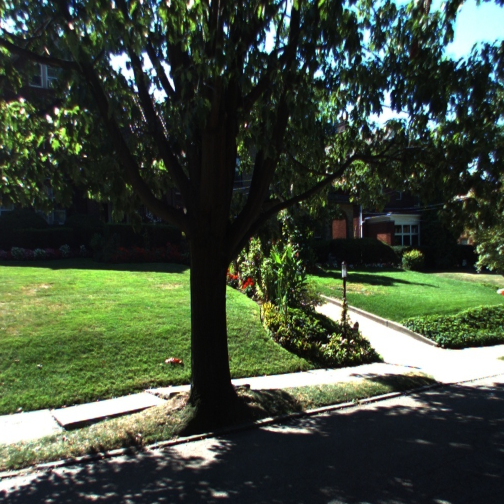} &
        \includegraphics[width=\imgwidth]{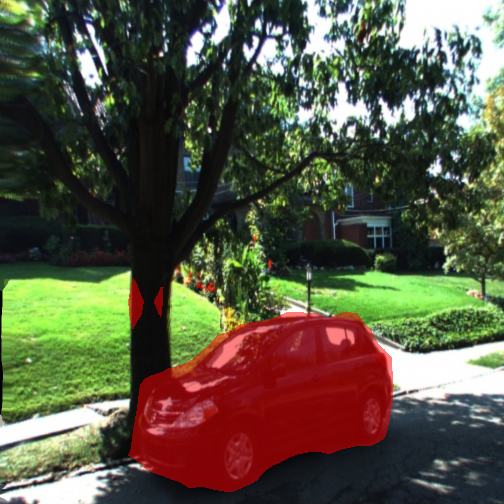} &
        \includegraphics[width=\imgwidth]{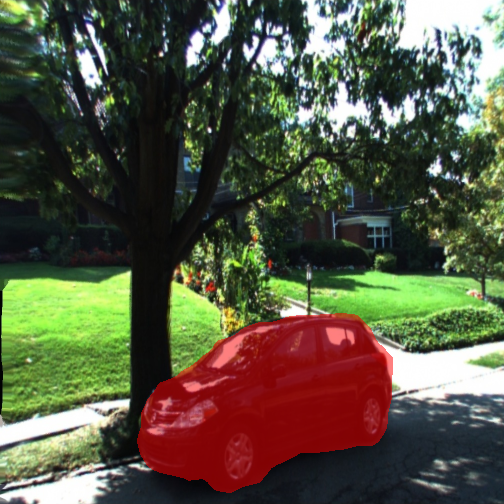} &
        \includegraphics[width=\imgwidth]{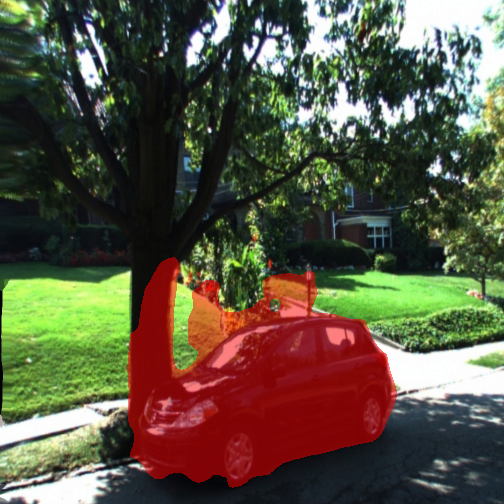} &
        \includegraphics[width=\imgwidth]{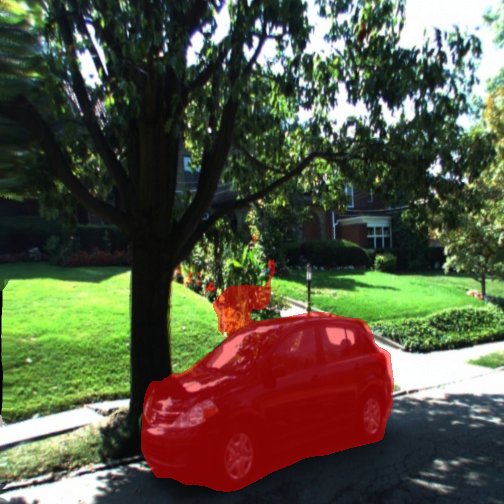} &
        \includegraphics[width=\imgwidth]{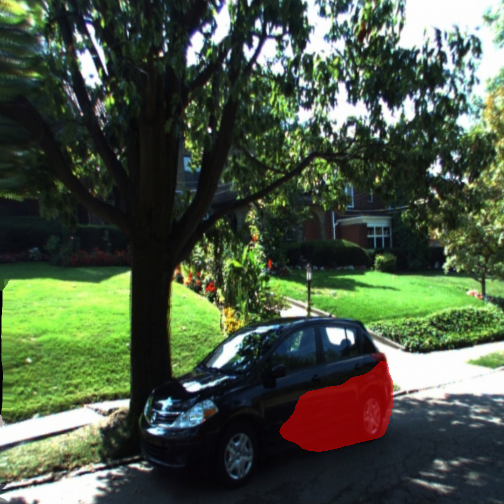} &
        \includegraphics[width=\imgwidth]{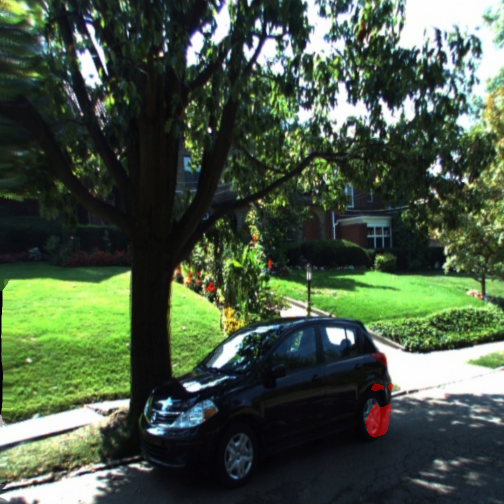} &
        \includegraphics[width=\imgwidth]{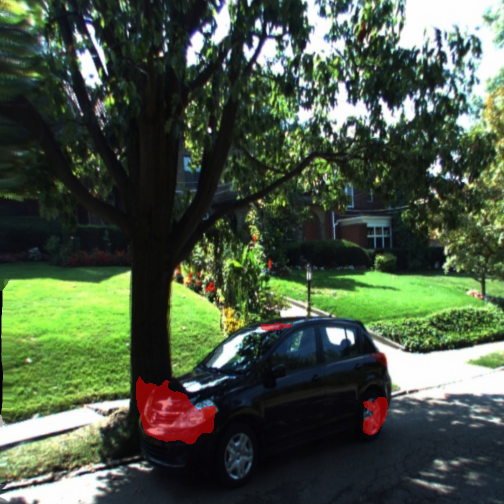} &
        \includegraphics[width=\imgwidth]{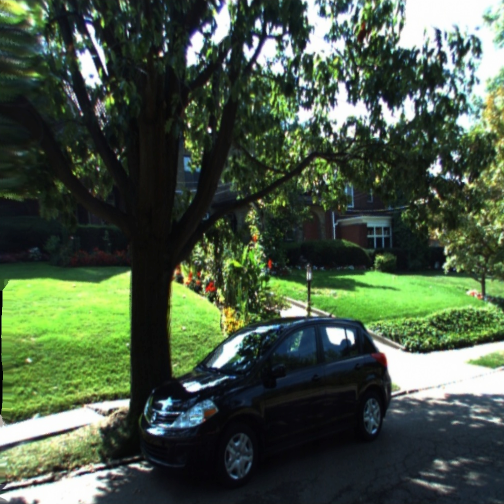} \\

        \midrule

        \includegraphics[width=\imgwidth]{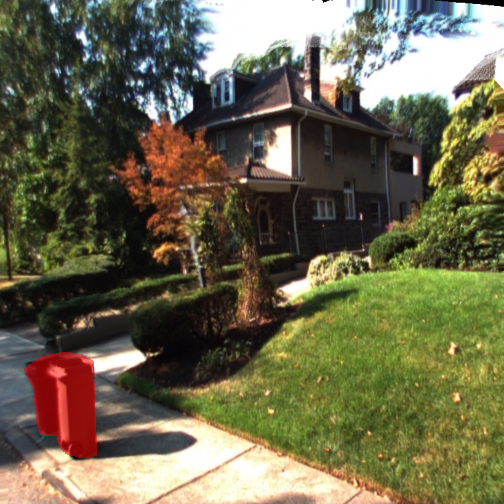} &
         &
        \includegraphics[width=\imgwidth]{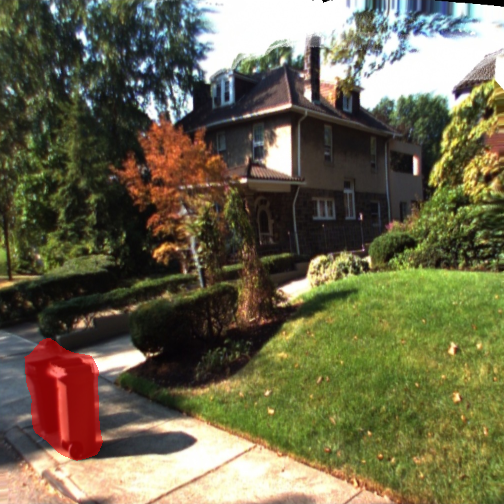} &
        \includegraphics[width=\imgwidth]{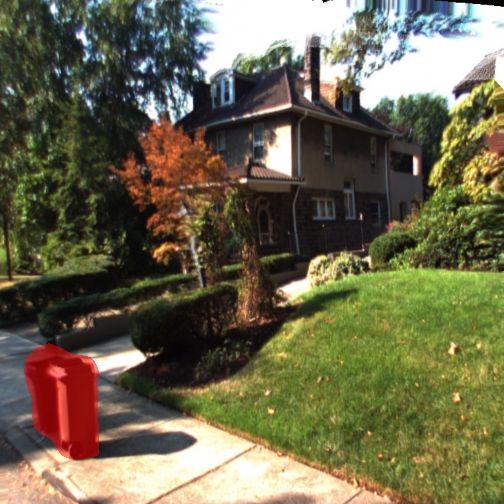} &
        \includegraphics[width=\imgwidth]{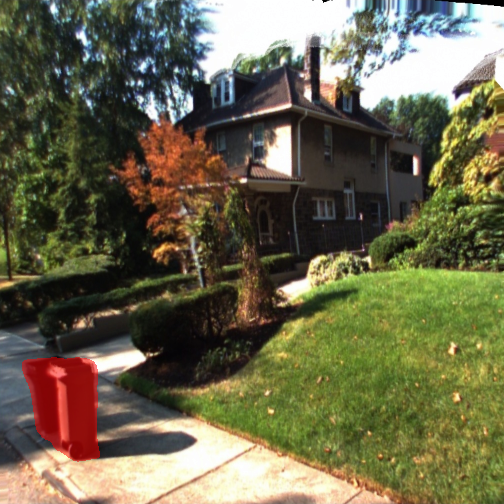} &
        \includegraphics[width=\imgwidth]{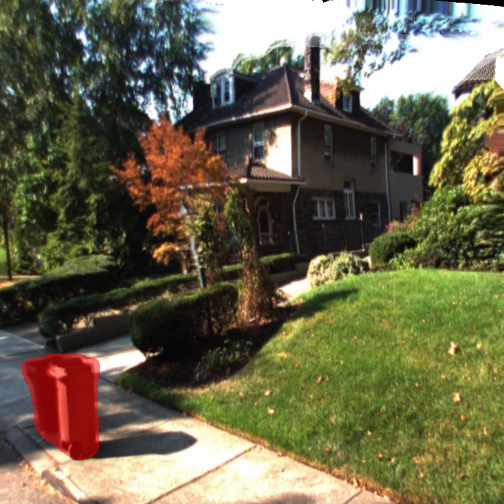} &
        \includegraphics[width=\imgwidth]{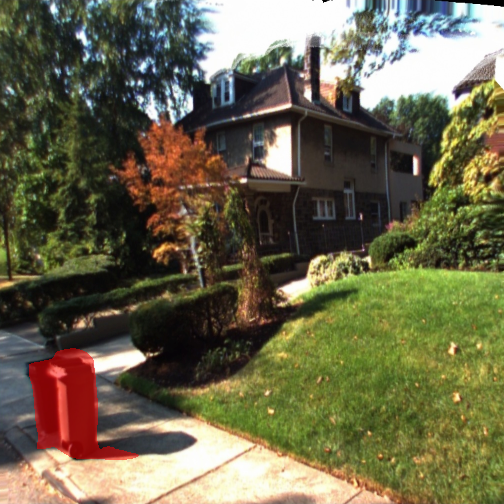} &
        \includegraphics[width=\imgwidth]{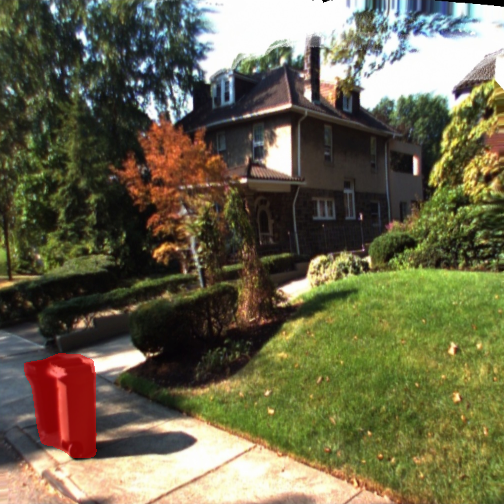} &
        \includegraphics[width=\imgwidth]{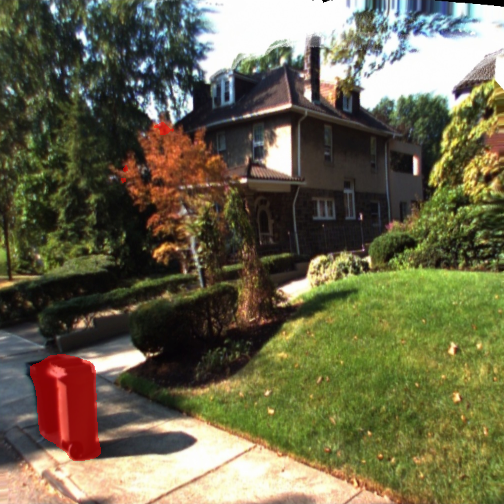} &
        \includegraphics[width=\imgwidth]{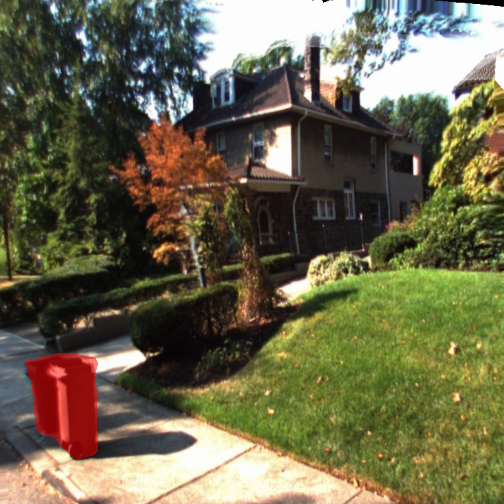} \\

        \includegraphics[width=\imgwidth]{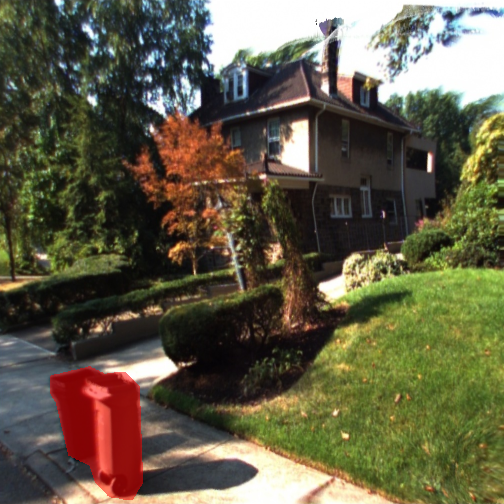} &
        \includegraphics[width=\imgwidth]{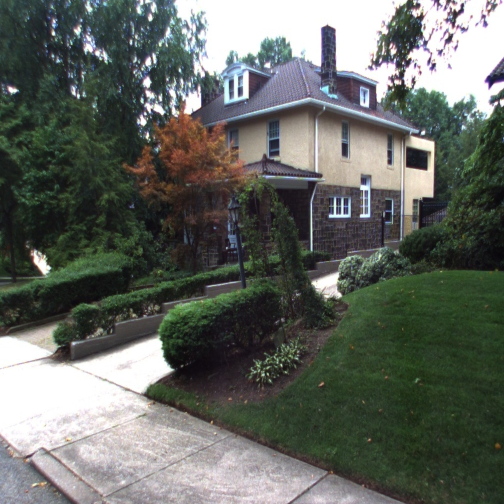} &
        \includegraphics[width=\imgwidth]{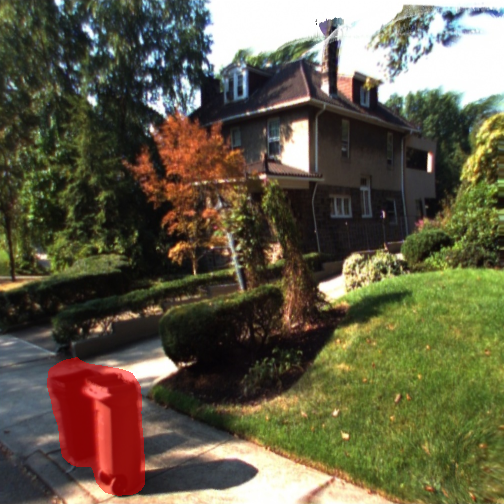} &
        \includegraphics[width=\imgwidth]{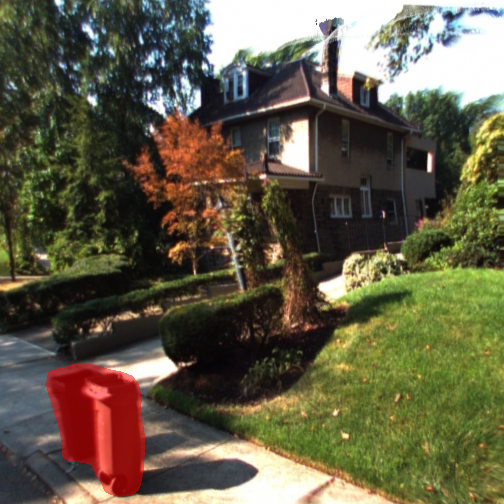} &
        \includegraphics[width=\imgwidth]{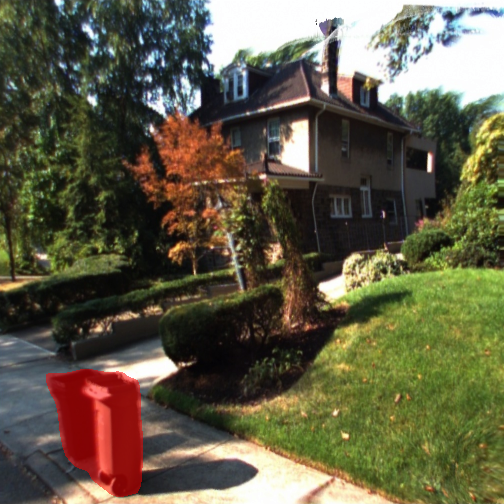} &
        \includegraphics[width=\imgwidth]{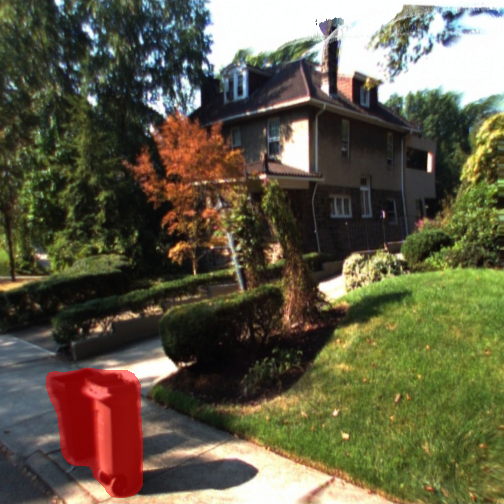} &
        \includegraphics[width=\imgwidth]{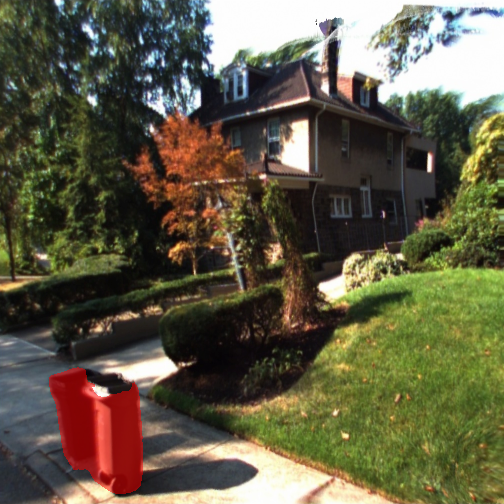} &
        \includegraphics[width=\imgwidth]{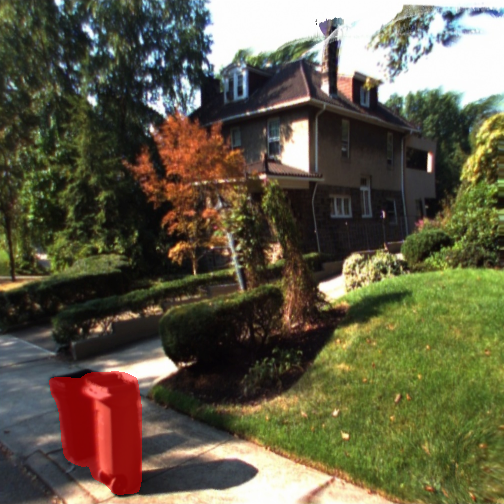} &
        \includegraphics[width=\imgwidth]{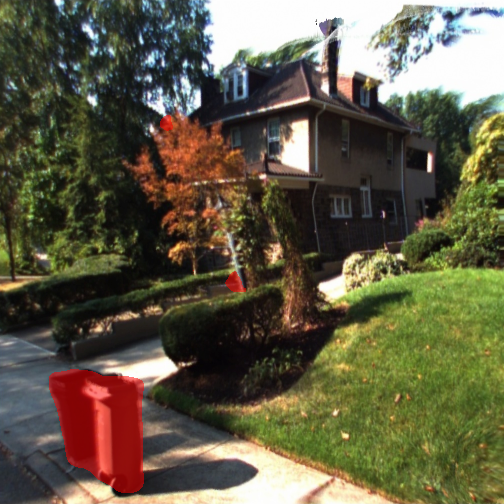} &
        \includegraphics[width=\imgwidth]{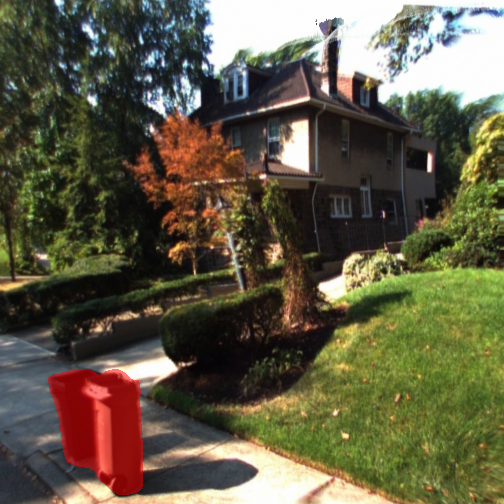} \\

        \includegraphics[width=\imgwidth]{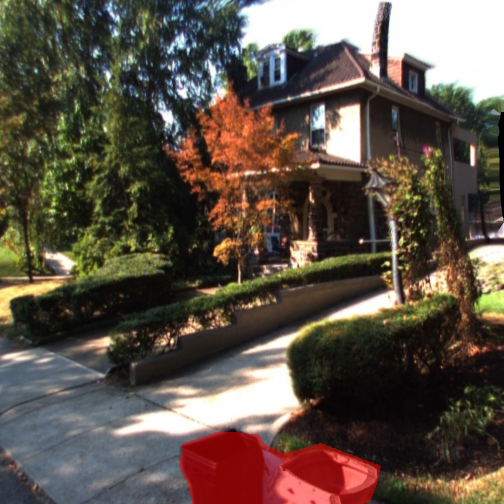} &
         &
        \includegraphics[width=\imgwidth]{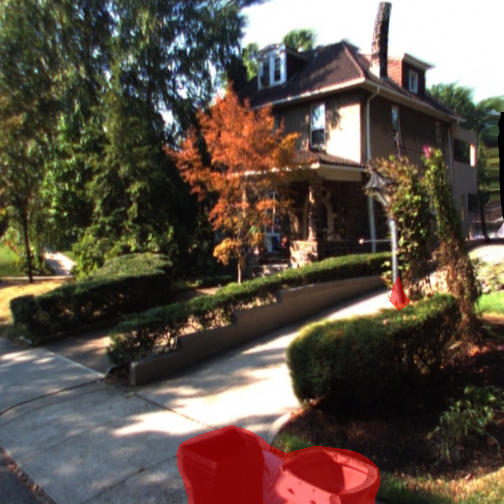} &
        \includegraphics[width=\imgwidth]{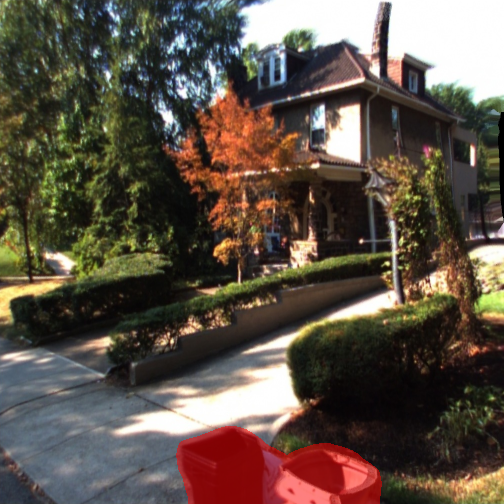} &
        \includegraphics[width=\imgwidth]{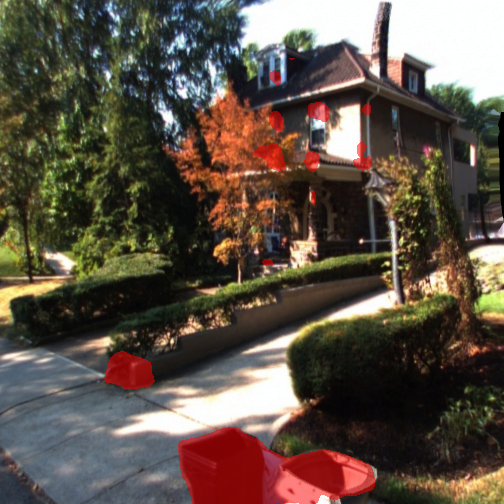} &
        \includegraphics[width=\imgwidth]{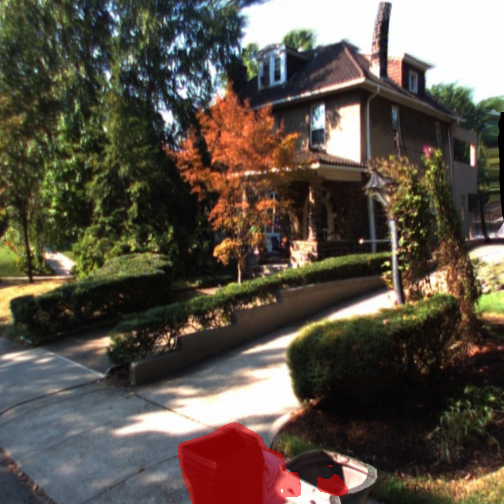} &
        \includegraphics[width=\imgwidth]{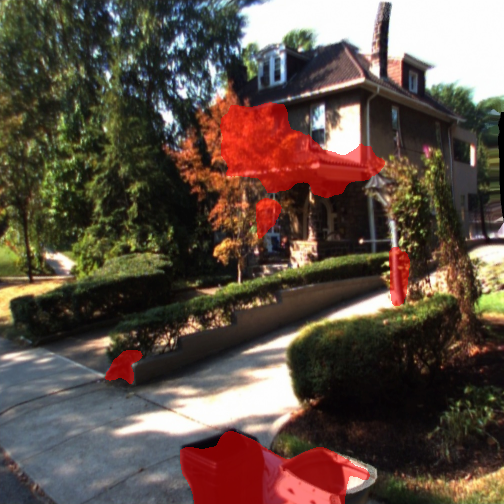} &
        \includegraphics[width=\imgwidth]{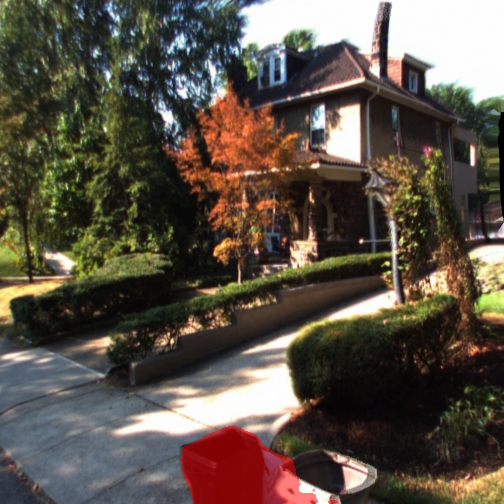} &
        \includegraphics[width=\imgwidth]{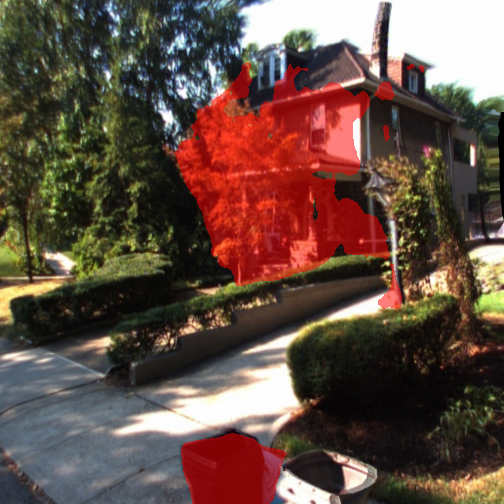} &
        \includegraphics[width=\imgwidth]{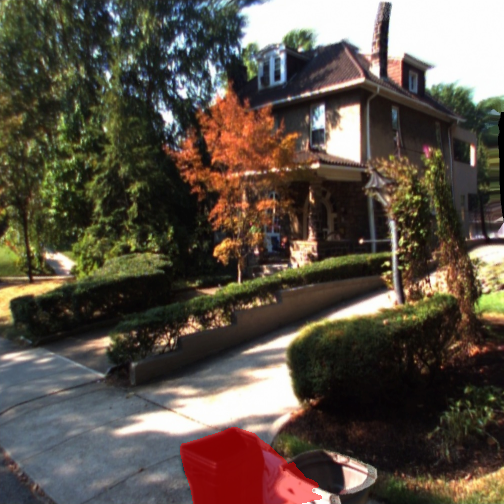} \\

        \bottomrule
    \end{tabular}
    \caption{\textbf{\textit{Qualitative Results:}} we visualize results from ``Aligned" of VL-CMU-CD in rows 2 and 5. The other rows are from ``Diff-2". The first scene compares the same $t_0$ image with a sequence of $t_1$ images, while the other compares the opposite.}
    \label{fig:cmu-cd}
\end{figure*}

\setlength{\tabcolsep}{6pt}

\begin{table*}[!t]

    \centering
    \caption{\textbf{\textit{F1-score after fine-tuning on PSCD:}} we report F1-scores of aligned/unaligned of VL-CMU-CD and PSCD to compare adaption ability with \cref{tab:table-backbone}.}
    \resizebox{\textwidth}{!}{
        \begin{tabular}{l | c | c c c c | c c c c | c}
                
            \toprule
            
            \multicolumn{1}{l|}{} &
            \multicolumn{1}{l|}{} & 
            \multicolumn{4}{c|}{Vl-CMU-CD~\cite{vl-cmu-cd}} &
            \multicolumn{4}{c|}{PSCD~\cite{sscdnet}} &
            \multicolumn{1}{c}{both} \\
            
            Method & Backbone &
            Aligned & Diff-1 & Diff-2 & Avg. & Aligned & Diff-1 & Diff-2 & Avg. & Avg.\\
            
            \midrule
            
            DR-TANet~\cite{drtanet} & Resnet-18 & 0.390 & 0.366 & 0.354 & 0.367 & 0.190 & 0.169 & 0.125 & 0.157 & 0.211 \\
            C-3PO~\cite{c3po} & Resnet-18 & 0.465 & 0.391 & 0.367 & 0.400 & 0.433 & 0.246 & 0.165 & 0.256 & 0.293 \\
            \textbf{ours} & Resnet-18 & 0.400 & 0.383 & 0.374 & 0.384 & 0.382 & 0.281 & \textbf{0.192} & 0.269 & 0.299 \\
            \textbf{ours} & DinoV2 (frozen) & \textbf{0.649} & \textbf{0.604} & \textbf{0.580} & \textbf{0.606} & \textbf{0.442} & \textbf{0.284} & 0.191 & \textbf{0.284} & \textbf{0.366} \\
            
            \bottomrule
            
        \end{tabular}
    }
    
    \label{tab:after-fine-tune}
\end{table*}

\begin{table*}[t]

    \centering
    \caption{\textbf{\textit{F1-score of Different Feature Comparator:}} we compare the results after replacing our cross-attention modules with feature comparators from baselines.}
    \resizebox{0.95\textwidth}{!}{
        \begin{tabular}{l | c c c c | c c c c | c}
            
            \toprule
            
            \multicolumn{1}{l|}{} &
            \multicolumn{4}{c|}{Vl-CMU-CD~\cite{vl-cmu-cd}} &
            \multicolumn{4}{c|}{PSCD~\cite{sscdnet}} &
            \multicolumn{1}{c}{both} \\
            
            Comparator &
            Aligned & Diff-1 & Diff-2 & Avg. & Aligned & Diff-1 & Diff-2 & Avg. & Avg.\\
            
            \midrule
            
            Co-Attention~\cite{cyws, wiles2021co} & 0.670 & 0.651 & 0.641 & 0.652 & 0.228 & 0.182 & 0.136 & 0.175 & 0.297\\

            Temporal Attention~\cite{drtanet} & 0.759 & 0.734 & 0.715 & 0.733 & 0.282 & 0.241 & 0.183 & 0.228 & 0.358 \\
            
            MTF~\cite{c3po} & 0.786 & 0.697 & 0.658 & 0.704 & 0.299 & 0.254 & 0.178 & 0.235 & 0.355 \\
            
            Cross Attention (\textbf{ours}) & \textbf{0.795} & \textbf{0.760} & \textbf{0.739} & \textbf{0.761} & \textbf{0.337} & \textbf{0.287} & \textbf{0.204} & \textbf{0.267} & \textbf{0.393} \\
            
            \bottomrule
            
        \end{tabular}
    }
    
    \label{tab:ablation}
\end{table*}

\begin{table}[htbp]

    \centering
    \caption{\textbf{\textit{Choice of Architecture:}} we compare different backbones with different cross-attention composition to specify our motivation of using the DinoV2 backbone and two cross-attentions.}
    \resizebox{0.5 \textwidth}{!}{
        \begin{tabular}{l | c | c | c | c}

            \toprule
            
            Method & Backbone & VL-CMU-CD~\cite{vl-cmu-cd} & PSCD~\cite{sscdnet} & Avg. \\
            
            \midrule

            2 CrossAttn & Resnet-18 & 0.687 & 0.097 & 0.257\\
            1 CrossAttn & DinoV2 & 0.762 & 0.326 & 0.444 \\            
            2 CrossAttn & DinoV2 & \textbf{0.795} & \textbf{0.337} & \textbf{0.461} \\
            \bottomrule
            
        \end{tabular}
    }
    \label{tab:table-arch-choice}
\end{table}

\subsection{Implementation Details}

We followed the training setting from C3PO~\cite{c3po}, using the Adam optimizer~\cite{adam}, 0.0001 initial learning rate, and the cosine learning-rate decay strategy. We use the weighted softmax cross-entropy loss function during training, and the weights for the change and unchanged classes are 0.975 and 0.025, respectively. The significant difference between the change and unchanged weight is because most change objects in the datasets only take a small fraction of a whole image. Regarding the training hardware, we used one NVIDIA A100 Tensor Core GPU to train with batch size 4.

\section{Results}

\subsection{Viewpoint Robustness}
\label{subsec:train-CMU}

We compare a series of baselines in \cref{tab:table-backbone} and report their respective performance on the aligned and unaligned VL-CMU-CD and PSCD datasets. For the TransCD~\cite{wang2021transcd} and C-3PO~\cite{c3po}, we evaluate results by their providing pre-trained weight. We trained DR-TANet~\cite{drtanet} and CDNet~\cite{sakurada2017dense} on our VL-CMU-CD training set, as they do not provide pre-trained weights. Meanwhile, we report CSCDNet~\cite{sscdnet} result from the paper of C3PO~\cite{c3po} for reference.

Comparing with the state-of-art C3PO~\cite{c3po}, we get a comparable result on the aligned VL-CMU-CD dataset. However, the performance gain increases drastically with the unaligned data, ``Diff-1" and ``Diff-2" of VL-CMU-CD, as the adjacent distance increases. Meanwhile, the PSCD column in \cref{tab:table-backbone} indicates the generalization ability because all methods are trained with aligned VL-CMU-CD data only. We infer that our F1-score of 0.337 of PSCD is attributed to the DinoV2 backbone by the results of replacing DinoV2 with Resnet-18 backbone in our architecture. Notably, this high performance is achieved without exposing the model to viewpoint variations/augmentations in the training set.

\subsection{Different Viewpoint Augmentation}
\label{subsec:diff-view-aug}

For the extensive study on how different viewpoints affect the performance, we append a training set augmented by affine transformation and a new training set ``Diff-1" into the original VL-CMU-CD training set, where affine transformation constitutes of random rotation within 15 degrees and translation within 50 pixels and the ``Diff-1" is the adjacent pair of the training set with the distance equal to 1. We retrain methods with these augmented datasets and evaluate by the VL-CMU-CD in \cref{tab:diff-view-augt}. All methods drop their performance on the ``Aligned" metric, but the difference between ``Aligned" and ``Diff-1" notably decreases after the different viewpoint augmentation. Consequently, our method remains the finest whether different viewpoint augmentation is applied.

We also report the affine transformation results in \cref{fig:fig-affine}. For translation evaluation, we translate the test set from 0 to 255 pixels and average the F1-score in four directions: right, left, up, and down. Moreover, we rotate the test set from 0 to 45 degrees and average the F1-score clockwise and counterclockwise. As a result, our method is the most robust on affine transformation among these baselines.

\subsection{Adapting to Unseen Data}
\label{subsec:ft-PSCD}

To analyze the ability of different methods to adapt to unseen data, we evaluate a few baselines after fine-tuning the VL-CMU-CD models using the PSCD dataset, as shown in \cref{tab:after-fine-tune}. Since all models are fine-tuned to adapt to the PSCD dataset, their performance retention on the base dataset (VL-CMU-CD) and performance growth on the fine-tuning dataset (PSCD) are indicators of how well a method can adapt to novel environments. Comparing \cref{tab:table-backbone} and \cref{tab:after-fine-tune}, the performance of baselines grows significantly on the PSCD dataset but only at the cost of a major drop in performance on the VL-CMU-CD. We infer that it is reasoned by the Resnet-18 backbone as our model with Resnet-18 backbone also suffers the same cost. On the other hand, our proposed method with DinoV2 backbone exhibits much better adaptation ability by comparing both VL-CMU-CD and PSCD.

\subsection{Qualitative Analyses}
\label{subsec:qual_analyses}

To understand how change masks are changed after the different viewpoint augmentation, we visualize both ``Aligned" and ``Diff-2" scenarios of the VL-CMU-CD in \cref{fig:cmu-cd}. There are two scenes in \cref{fig:cmu-cd}, where the first one is to detect changes with a ``$t_0$" image and a sequence of ``$t_1$" images, and the other one is to detect changes with a sequence of ``$t_0$" images and a ``$t_1$" image. We can observe that all methods reduce false positives on ``Diff-2" cases after training with different viewpoint augmentation.

\subsection{Ablation Study: Comparing the Comparator}
\label{subsec:ablation}

We have compared how different backbones affect the performance in our architecture. We further compare different feature comparators in \cref{tab:ablation}. Specifically, we take the Merge Temporal Feature (MTF) module from C3PO \cite{c3po}, co-attention from \cite{cyws, wiles2021co}, and Temporal Attention from DR-TANet\cite{drtanet} to replace with cross-attention modules in \cref{fig:architecture}. Hence, all feature comparators of baselines utilize the exact same dense features from DinoV2, achieving a fair comparison of feature comparators. As a result, all baselines achieve better performance on ``PSCD" metrics comparing to \cref{tab:table-backbone} as the DinoV2 backbone brings generalization ability. However, peak performance is only achieved when this backbone is combined with the cross-attention-based comparator, as observed in the last row (ours) of \cref{tab:ablation}.

\subsection{Ablation Study: Choices of Architecture}

We report results about changing DinoV2 to Resnet-18 and reduce two to one cross-attention in the \cref{tab:table-arch-choice}. We can tell that DinoV2 is a superior backbone to Resnet-18, and two cross-attentions significantly leverage the performance.

\section{Conclusion}
\label{sec: conclusion}
We introduced a novel scene change detection method leveraging DINOv2's robust feature extraction and cross-attention modules to handle challenges like lighting, weather, and viewpoint differences. Our approach demonstrated significant improvements in F1-score on the VL-CMU-CD and PSCD datasets, showing better generalization and robustness against photometric and geometric variations.

By effectively managing correspondences between image pairs, our method outperformed existing approaches and demonstrated strong performance in scenarios involving geometric changes. This robust solution is applicable in autonomous driving, urban planning, environmental monitoring, and surveillance. Future work will focus on further model enhancements and the incorporation of additional contextual information to improve detection accuracy.

\section{Acknowledgment}
This work was supported with supercomputing resources provided by the Phoenix HPC service at the University of Adelaide.
{\small
\bibliographystyle{abbrv}
\bibliography{egbib}
}

\end{document}